\documentclass[letterpaper, 10 pt, conference]{ieeeconf}  

\IEEEoverridecommandlockouts                              

\title{\LARGE \bf
Interpretable Multimodal Gesture Recognition for Drone and Mobile Robot Teleoperation via Log-Likelihood Ratio Fusion
}

\author{Seungyeol Baek$^{1}$, Jaspreet Singh$^{2}$, Lala Ray$^{2,3}$, Hymalai Bello$^{3}$, Paul Lukowicz$^{2,3}$, and Sungho Suh$^{1}$
\thanks{$^{1}$S. Baek and S. Suh are with the Department of Artificial Intelligence, Korea University, Seoul, Republic of Korea
        {\tt\small \{mbaek01,sungho\_suh\}@korea.ac.kr}}%
\thanks{$^{2}$ J. Singh, P. Lukowicz are with the Department of Computer Science, RPTU Kaisersalutern-Landau, Kaiserslautern, Germany}
\thanks{$^{3}$ L.S.S. Ray, H. Bello, and P. Lukowicz are with the Embedded Intelligence, German Research Center for Artificial Intelligence (DFKI), Kaiserslautern, Germany
        {\tt\small \{lala\_shakti\_swarup.ray, hymalai.bello, paul.lukowicz\}@dfki.de}}%
\thanks{Code and additional materials: 
\url{https://github.com/mbaek01/Gesture-Recognition-for-Drone-Control}.}%
}
\usepackage{amssymb}
\usepackage{amsmath}
\usepackage{graphicx}
\usepackage{tabularx}
\usepackage{booktabs}
\usepackage{multirow}
\usepackage{bm}
\usepackage[hidelinks]{hyperref}
\usepackage[capitalise]{cleveref}
\usepackage{subcaption} 
\usepackage{flushend}

\begin{document}

\maketitle
\thispagestyle{empty}
\pagestyle{empty}

\begin{abstract}

Human operators are still frequently exposed to hazardous environments such as disaster zones and industrial facilities, where intuitive and reliable teleoperation of mobile robots and Unmanned Aerial Vehicles (UAVs) is essential. In this context, hands-free teleoperation enhances operator mobility and situational awareness, thereby improving safety in hazardous environments. While vision-based gesture recognition has been explored as one method for hands-free teleoperation, its performance often deteriorates under occlusions, lighting variations, and cluttered backgrounds, limiting its applicability in real-world operations. To overcome these limitations, we propose a multimodal gesture recognition framework that integrates inertial data (accelerometer, gyroscope, and orientation) from Apple Watches on both wrists with capacitive sensing signals from custom gloves. We design a late fusion strategy based on the log-likelihood ratio (LLR), which not only enhances recognition performance but also provides interpretability by quantifying modality-specific contributions. To support this research, we introduce a new dataset of 20 distinct gestures inspired by aircraft marshalling signals, comprising synchronized RGB video, IMU, and capacitive sensor data. Experimental results demonstrate that our framework achieves performance comparable to a state-of-the-art vision-based baseline while significantly reducing computational cost, model size, and training time, making it well suited for real-time robot control. We therefore underscore the potential of sensor-based multimodal fusion as a robust and interpretable solution for gesture-driven mobile robot and drone teleoperation.

\end{abstract}

\begin{keywords}
Multimodal Gesture Recognition, Wearable Sensors, Log-Likelihood Ratio Fusion, Drone and Mobile Robot Teleoperation
\end{keywords}

\section{Introduction}

Despite rapid advances in automation, many hazardous tasks in domains such as industrial manufacturing, construction, and emergency response are still performed directly by human workers. Firefighters, rescue teams, and industrial operators remain regularly exposed to extreme heat, toxic gases, unstable structures, and other life-threatening conditions. To reduce human risk, mobile robots and Unmanned Aerial Vehicles (UAVs) are increasingly deployed as substitutes or assistants in such environments, supporting tasks such as surveillance in confined industrial facilities \cite{azeta2019android,shanti2022real}, industrial automation \cite{wang2021training}, delivery of supplies in disaster zones \cite{euchi2021drones}, and emergency rescue operations \cite{rubio2019review, khan2022emerging}. Although these systems are expected to operate where humans cannot remain safely or efficiently, they often still rely on effective human teleoperation to ensure reliable task execution.

Teleoperation methods provide operators with a means to control robots remotely, bridging the gap between human decision-making and robotic actuation. Conventional approaches rely on physical controllers such as joysticks and consoles \cite{fong2001vehicle,cruz2022mixed}, which remain the standard in many industrial and defense applications. However, these devices restrict operator mobility and demand continuous manual engagement, limiting their suitability in highly dynamic or dangerous scenarios. As an alternative, gesture-based control has gained attention for its intuitive, hands-free, and natural interaction paradigm \cite{de2022gestures,ma2017hand}. By leveraging human body movements, operators can issue commands while maintaining situational awareness and physical flexibility—critical requirements in real-world deployments.

\begin{figure}[tb]
    \centering
    \includegraphics[width=\columnwidth]{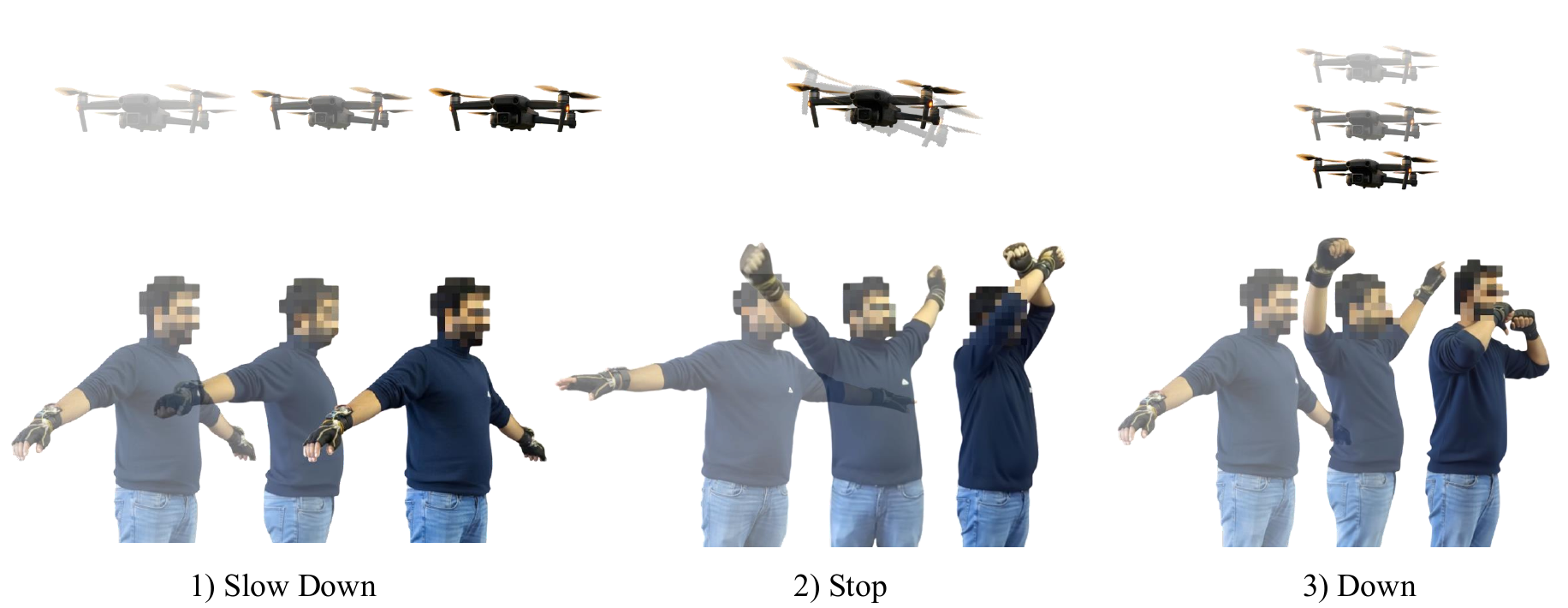}
    \caption{An example of drone control using the proposed sensor-based gesture recognition.}
    \label{fig:overview}
    \vspace{-5mm}
\end{figure}

A large body of work has explored vision-based gesture recognition, driven by the maturity of computer vision methods and the availability of visual sensors. Yet vision-based systems remain highly sensitive to environmental conditions: occlusions, poor lighting, camera placement, and background clutter often degrade recognition performance \cite{zhu2013vision}. In real operational settings such as smoke-filled disaster zones, dark industrial environments, or outdoor missions with variable lighting, these limitations hinder reliability. Consequently, vision-only systems struggle to meet the robustness required for mission-critical human–robot interaction.

Sensor-based gesture recognition using wearable devices offers a promising solution \cite{suh2023worker}.
By capturing motion and physiological signals directly from the operator’s body, wearable sensors provide robustness against environmental disturbances, enabling reliable operation where vision-based methods fail. However, single-modality systems often fall short of capturing the full richness of human gestures. Furthermore, many multimodal fusion techniques treat modality integration as a black box, offering little transparency regarding how different sensors contribute to recognition, a serious concern for safety-critical robotic control.

To address these challenges, we propose a multimodal gesture recognition framework tailored for mobile robots and drones. Our framework integrates accelerometer, gyroscope, and orientation data from Apple Watches worn on both wrists, together with capacitive sensing from our custom gloves \cite{bello2023captainglove}. These heterogeneous signals are combined through a log-likelihood ratio (LLR)–based late fusion strategy, which not only improves recognition accuracy but also provides interpretability by quantifying modality-specific contributions. This property is essential in robotic applications where operators and system designers must understand which signals drive the model’s decisions.

In support of this framework, we introduce a novel dataset of 20 distinct gestures inspired by aircraft marshalling signals \cite{choi2008visual}, a standardized communication protocol originally developed for ground–aircraft coordination. The dataset captures synchronized RGB video, wearable inertial data, and capacitive signals across both hands, making it a unique resource for evaluating multimodal gesture recognition models. Beyond drone control, the dataset and framework generalize to other mobile robot interaction scenarios, supporting broader adoption of gesture-based control systems.

Finally, we evaluate multiple fusion strategies, including LLR and self-attention, and conduct interpretability analysis and ablation studies to assess the importance of each modality. Comparisons with the state-of-the-art (SoTA) vision-based method PoseConv3D \cite{duan2022revisiting} show that our sensor-based multimodal approach outperforms the vision-based method while reducing computational cost, model size, and training time, making it suitable for real-time robotic deployment. 

Our contributions are summarized as follows:
\begin{itemize}
    \item We propose a deep learning framework that integrates heterogeneous wearable sensor modalities through a log-likelihood ratio (LLR)–based late fusion strategy, combining accuracy with interpretability.
    \item We provide interpretability analyses of fusion methods, leveraging LLR values and attention weights to quantify modality-specific contributions to recognition.
    \item We introduce a novel multimodal dataset for mobile robot and drone control, comprising 20 distinct gestures inspired by aircraft marshalling signals, with synchronized RGB video, inertial, and capacitive sensor data.
    \item We conduct ablation studies to evaluate robustness under missing-modality conditions and to highlight the role of different sensors.
    \item We show that sensor-based multimodal approaches achieve performance comparable to SoTA vision-based methods, while reducing computational cost, model size, and training time for practical robotic deployment.
\end{itemize}

\section{Related Works}
\subsection{Applications of Mobile Robotics}
Mobile robots and UAVs are increasingly deployed in safety-critical domains such as disaster response, industrial inspection, and logistics \cite{moniruzzaman2022teleoperation,opiyo2021review}.
For example, mobile robots are used for monitoring and surveying hazardous industrial facilities, detecting safety risks at construction sites, and enabling smart manufacturing through robot-assisted production. 
UAVs and drones are also being explored for contactless delivery of medical supplies, disaster detection, and emergency response \cite{chiou2021mixed}.

Despite advances in autonomy, many of these tasks still require reliable human oversight and control to ensure safe and effective operation \cite{zhou2022teleman}. This makes intuitive and robust teleoperation methods critical for real-world deployment.

\subsection{Control Methods for Mobile Robotics}
Traditional robot teleoperation relies on physical interfaces such as joysticks, keyboards, and specialized control consoles, which remain standard in industrial applications. While these devices provide precision, they limit operator mobility and require continuous manual engagement, reducing effectiveness in dynamic and hazardous environments \cite{esaki2024immersive,lee2023investigating}. 
Alternative approaches have explored mixed reality interfaces for more flexible robot control \cite{cruz2024analysis, adekoya2025horus}. However, these systems often demand cumbersome infrastructure and remain difficult to scale in the field. Overall, existing control methods highlight a trade-off between precision and operator flexibility, motivating the search for more natural, hands-free interaction paradigms.

\begin{figure}[!t] 
    \centering 
    \begin{subfigure}[b]{0.48\columnwidth} 
        \centering
        \includegraphics[width=\linewidth, height=3.5cm]{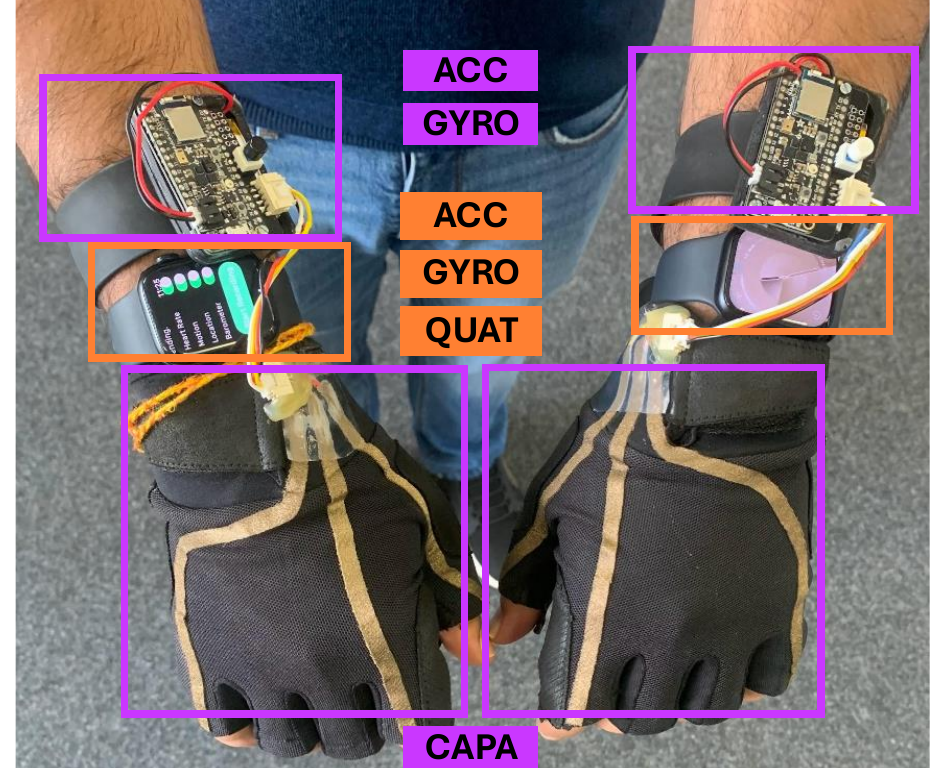} 
    \vspace{-5mm}
    \caption{Textile sensing gloves and Apple Watch devices}
    \label{fig:all_devices}
    \end{subfigure}
    \hfill
    \begin{subfigure}[b]{0.48\columnwidth} 
        \centering
        \includegraphics[width=\linewidth, height=3.5cm]{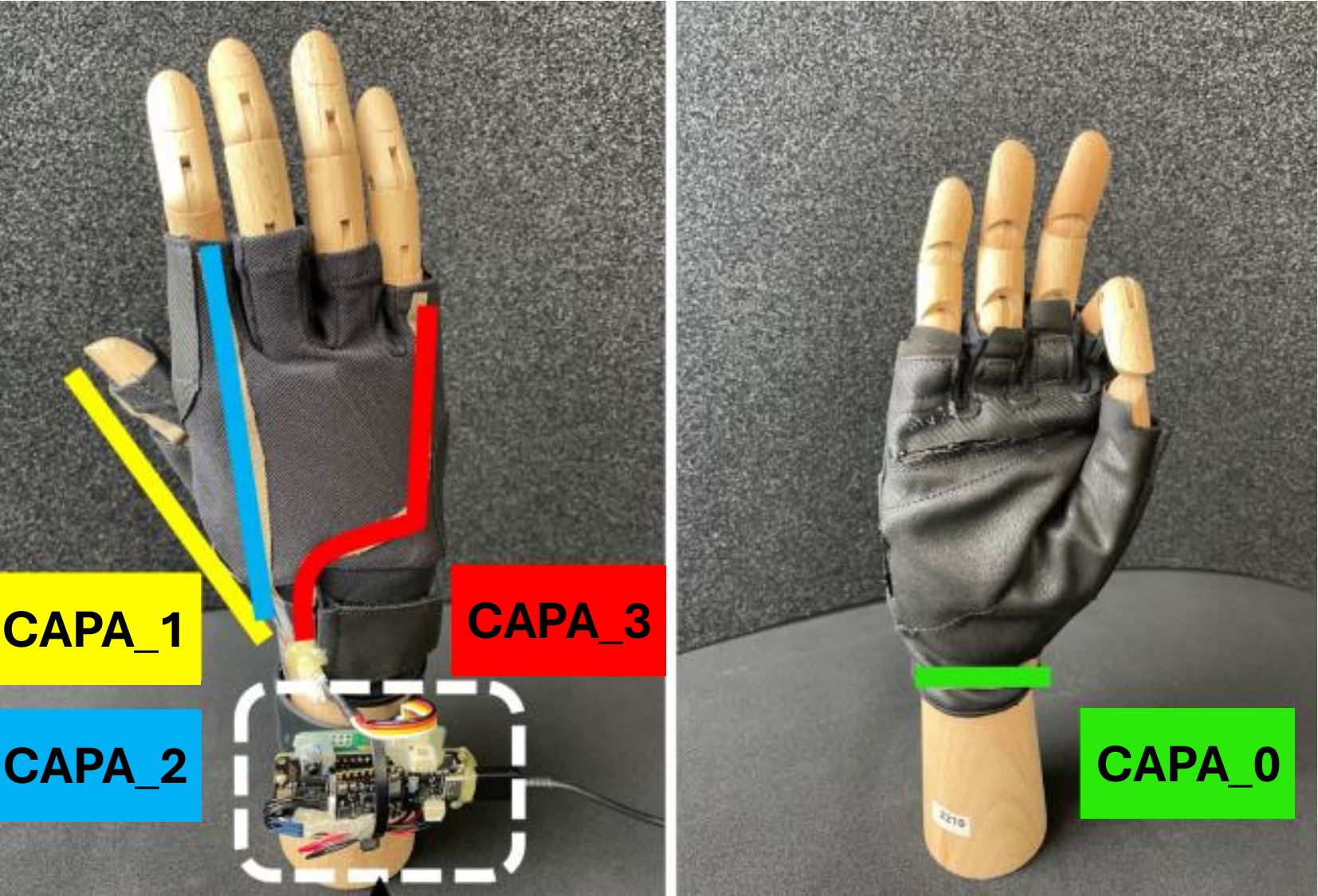}
        \caption{Capacitive sensor channels on the gloves \cite{bello2023captainglove}}
        \label{fig:chnl_glove}
    \end{subfigure}
    \caption{Hardware setup with textile sensing gloves and Apple Watch devices on both hands is shown in \cref{fig:all_devices}. Purple outlines mark capacitive sensors (bottom) and IMU sensors (top), while orange outlines mark Apple Watch devices on the wrist used for IMU sensing and quaternion orientation.}
    \vspace{-5mm}
    \label{fig:hardware}
\end{figure}

\subsection{Gesture Recognition as a Control Method}
Gesture recognition has been widely studied as a natural interface for human–computer interaction, enabling intuitive control in domains such as augmented and virtual reality \cite{di2024toward}, gaming \cite{sharma2024hand}, and smart environments \cite{panagiotou2025multidisciplinary}. Vision-based approaches have shown strong performance in controlled laboratory settings, but their reliability degrades under occlusion, variable lighting, and cluttered backgrounds \cite{qi2024computer}. To address these limitations, wearable sensor-based methods have gained attention, leveraging accelerometers, gyroscopes, and capacitive sensing to directly capture body movements \cite{bello2023captainglove}. These approaches have demonstrated robustness in human activity recognition tasks, where environmental variability poses significant challenges. More recent work has focused on multimodal fusion to exploit the complementary strengths of different sensors, though many fusion strategies still operate as black boxes, offering little interpretability of modality contributions \cite{ray2024har}.

While these techniques were originally developed for broader HCI and wearable computing applications, their advantages naturally extend to human–robot interaction. Gesture-based teleoperation of UAVs and multi-robot systems illustrates the potential of adapting these recognition methods to safety-critical robotic contexts, where robustness, efficiency, and interpretability are essential.

\section{Methodology}

\begin{figure}[tb]
    \centering
    \includegraphics[width=\columnwidth]{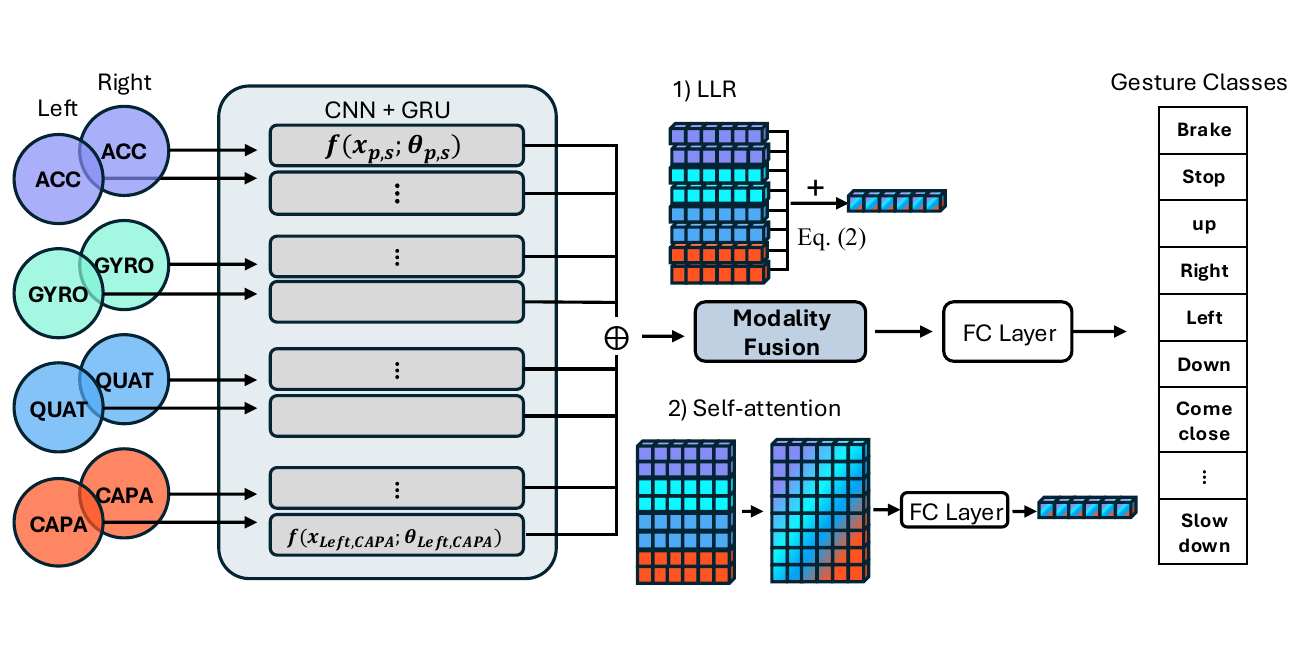}
    \caption{Framework architecture for sensor-based gesture recognition on our dataset. Each sensor modality is processed individually through convolutional and temporal feature extraction layers, and the resulting modality-specific features are fused using either log-likelihood ratio (LLR) fusion \cref{eq:llr_fusion} or self-attention fusion for classification.}
    \label{fig:model}
    \vspace{-5mm}
\end{figure}

\subsection{Hardware Setting} 

The dataset was collected using a multimodal wearable setup consisting of textile sensing gloves, Apple Watch devices, and an RGB camera (\cref{fig:hardware}). The gloves, adapted from a previously developed design \cite{bello2023captainglove}, integrate stretchable textile electrodes and an inertial measurement unit (IMU) into lightweight sports gloves. Four capacitive channels were distributed across the fingers and wrist, connected to an FDC2214 capacitance-to-digital converter, with signals sampled at approximately 50 Hz. Each glove included a wrist-mounted IMU to capture acceleration and angular velocity.

To complement the glove-based signals, Apple Watch devices (Series 7) were worn on both wrists, providing accelerometer, gyroscope, and orientation data at approximately 100 Hz. A ZED Mini stereo camera recorded synchronized RGB videos at 30 fps, ensuring alignment between visual and wearable modalities. This combined setup—capacitive glove electrodes, inertial wrist sensors, Apple Watches, and RGB video—enables comprehensive multimodal capture of hand and arm gestures for drone and robot teleoperation.

\subsection{Network Architecture}
The framework for multi-sensor gesture recognition processes each sensor modality individually to obtain its feature representation. \cref{fig:model} provides an overview of the entire network architecture, where we define the set of modality pairs as $\mathcal{M} = \{(p,s) \mid p \in \{\text{Left}, \text{Right}\}, s \in \{\text{ACC}, \text{GYRO}, \text{QUAT}, \text{CAPA}\}\}$, with ACC, GYRO, QUAT, and CAPA denoting the accelerometer, gyroscope, quaternion orientation, and capacitive sensor, respectively. The input sensor data are represented as $X_{p,s} \in \mathbb{R}^{T_{p,s} \times C_{p,s}}$, where $T_{p,s}$ denotes the temporal sliding window size and $C_{p,s}$ the number of channels for modality $(p,s)$. Each input sequence is processed independently by a feature extraction pipeline comprising convolutional and temporal layers to produce latent representations. These extracted features are subsequently fused using one of two approaches: log-likelihood ratio (LLR) fusion or self-attention fusion.

\subsubsection{Feature Extraction per Modality}
Feature extraction for each modality is carried out in two stages: an initial convolutional subnet for local feature extraction, followed by a temporal subnet based on gated recurrent units (GRUs) \cite{chung2014empirical}. Inspired by human activity recognition (HAR) models such as DeepConvLSTM \cite{ordonez2016deep} and TinyHAR \cite{zhou2022tinyhar}, we employ four convolutional subnets for initial feature extraction, one per modality. Each subnet consists of a 1D convolution layer applied along the temporal axis, followed by ReLU activations and batch normalization. The output shape of the convolutional subnet $\mathbb{R}^{T^*_{p,s} \times D}$, where $T^*_{p,s}$ indicates the reduced temporal dimension and $D$ the output channel dimension.

The extracted features from each modality are then passed independently through an attention-based GRU subnet to capture temporal dependencies. The subnet consists of a two-layer GRU that outputs hidden states at every timestep. Following prior work \cite{ma2019attnsense, zhou2022tinyhar}, rather than relying solely on the final hidden state as the temporal representation, we apply a self-attention mechanism to compute a weighted sum of all hidden states as a global temporal context representation. This global temporal representation is multiplied by a trainable gating parameter, allowing the model to learn whether to emphasize or disregard the global representation, and is then added to the hidden state of the last timestep \cite{zhou2022tinyhar}. Therefore, the output of this temporal subnet has shape $\mathbb{R}^D$.

\subsubsection{Late Fusion Approaches (LLR and Self-Attention)}
We then apply either LLR or self-attention fusion to combine the per-modality features. In the LLR approach, the likelihood that a feature from a given modality belongs to a specific class is first computed. Denoting the per-modality feature as $h_{p,s} \in \mathbb{R}^D$, the predicted softmax output for the modality $(p,s)$ is $\hat{y}_{p,s}$, defined over a set of $N$ classes $\{C_i,..., C_n\}$. The estimated probability that $\hat{y}_{p,s}$ belongs to class $C_i$, denoted  $\hat{p}(\hat{y}_{p,s} \in C_i)$, is used to compute the LLR for that class as: 

\begin{equation}
\begin{split}
LLR_{p,s}(\hat{y}_{p,s} \in C_{i}) 
&= \ln \left( 
   \frac{\hat{p}(\hat{y}_{p,s} \in C_{i})}
        {\hat{p}(\hat{y}_{p,s} \notin C_{i})} 
   \right) \\
&= \ln \left( 
   \frac{\hat{p}(\hat{y}_{p,s} \in C_{i})}
        {\sum_{\substack{j=1 \\ j \neq i}}^{N} 
         \hat{p}(\hat{y}_{p,s} \in C_{j})} 
   \right) 
\end{split}
\label{eq:llr}
\end{equation}

The computed LLRs from all modality pairs form a tensor of shape 
$\mathbb{R}^{|\mathcal{M}| \times D}$, where $|\mathcal{M}|=8$ is the number of modality pairs. 
Summing over all modality pairs $(p,s) \in \mathcal{M}$ yields the fused representation of shape $\mathbb{R}^D$:

\begin{equation}
LLR_{\text{fused}} = \sum_{(p,s) \in \mathcal{M}} LLR_{p,s}.
\label{eq:llr_fusion}
\end{equation}

The self-attention fusion module operates on the concatenated features from all modality pairs, represented as a matrix of shape $\mathbb{R}^{|\mathcal{M}| \times D}$. Each row corresponds to the GRU-encoded representation of one modality pair, with no sequence dimension. Inspired by \cite{zhou2022tinyhar, abedin2021attend, vaswani2017attention}, we apply a scaled dot-product self-attention mechanism across the modality dimension to model inter-modality dependencies. The self-attention mechanism assigns relative importance to each modality by computing its similarity with all other modalities; these similarity scores are normalized into attention weights that determine how strongly each modality aggregates information from the others. Using these weights, each modality feature integrates information from the other modalities to produce a fused representation of shape $\mathbb{R}^{|\mathcal{M}| \times D}$. Finally, a fully connected layer reduces the modality dimension, mapping the fused representation from $\mathbb{R}^{|\mathcal{M}| \times D}$ to a single vector in $\mathbb{R}^D$. 

The outputs of the LLR and self-attention fusion modules therefore share the same shape, $\mathbb{R}^D$, which is further projected to $\mathbb{R}^N$ by an additional fully connected layer for final classification. To preserve a direct and interpretable relationship between the learned features and the final class predictions, no nonlinear activation is applied to the fusion outputs \cite{ribeiro2016should}.

\subsection{Model Interpretability}

We focus on post-hoc interpretability at the fusion stage, examining how LLR values and attention weights reflect modality contributions. Each fusion method provides interpretable signals of modality influence. Pre-fusion LLR values indicate each modality's relative contribution to the class prediction. Similarly, attention weights are examined to identify modality interactions indicated by high weights. Although attention weights do not directly quantify contributions, they provide insight into how strongly a query modality attends to others when forming a prediction \cite{vaswani2017attention, serrano2019attention}. We visualize these attention weights as heatmaps for inter-modality dependencies, alongside the per-modality LLR contributions. Additionally, we conduct an ablation study—evaluating performance by excluding sensors and testing different combinations—to further assess modality contributions.

\section{Experimental Setting}

\subsection{Gesture Definition}
\begin{figure}[tb]
    \centering
    \includegraphics[width=\columnwidth]{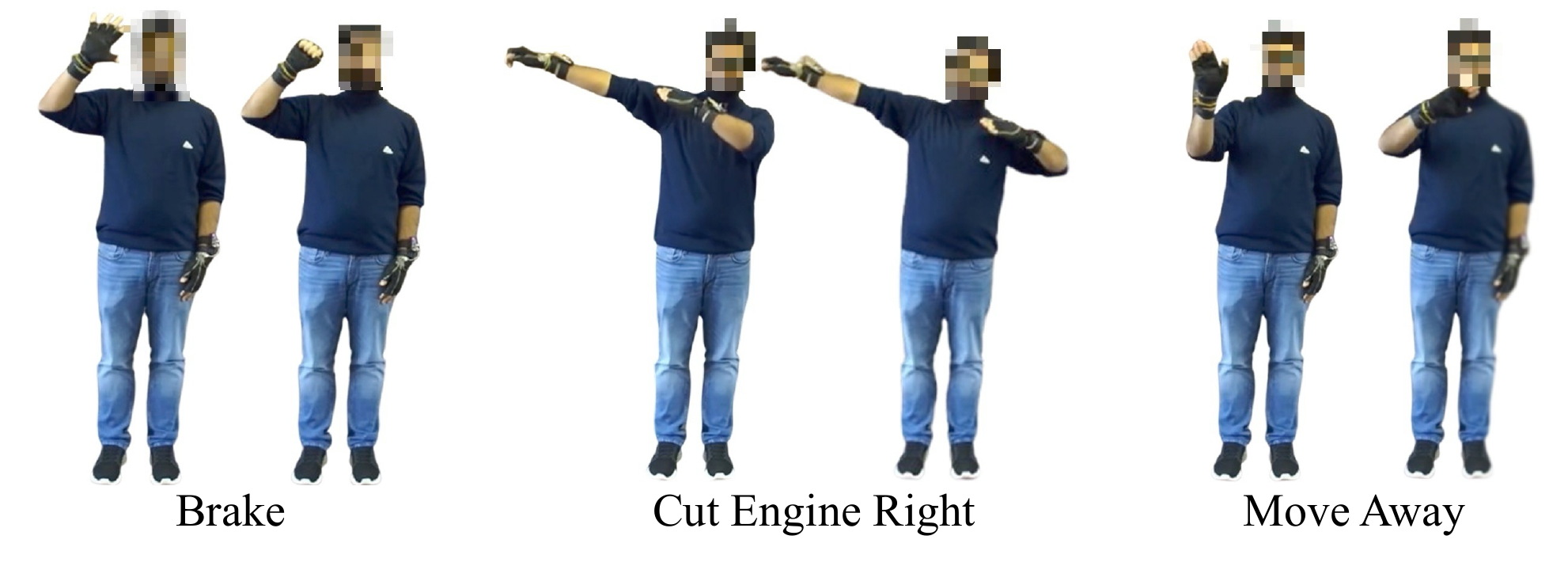}
    \caption{Example gesture classes are illustrated, including 'brake,' 'brake fire right,' and 'move away.' Each gesture is represented by two sequential image frames, displaying a non-mirrored, first-person perspective.}
    \vspace{-5mm}
    \label{fig:gesture}
\end{figure}

We define 20 distinct gesture classes, inspired by aircraft marshalling signals \cite{choi2008visual}. The gesture set, along with the corresponding label map ID used for classification, is summarized in \ref{tab:gesture_mapping}. Gestures were selected for their distinctive hand and arm motions to leverage the gloves’ textile capacitive sensors and the wrist-mounted IMU sensors. They were also designed for visibility and clarity across diverse environmental conditions. Example gestures are shown in \cref{fig:gesture}. While the gestures are well-suited and intuitive for aerial vehicles such as unmanned aerial vehicles (UAVs) due to their similarity in movement patterns to aircraft marshalling signals \cite{choi2008visual}, their general control patterns are applicable to a broader range of mobile robotic platforms. A comprehensive video of the defined gestures will be included as the accompanying video.

\begin{table}[t] \caption{Gesture Label Mapping} 
\label{tab:gesture_mapping} \centering 
\begin{tabularx}{\columnwidth}{Xc} \toprule \textbf{Gesture Label} & \textbf{ID} \\ \midrule Brake & 0 \\ Brake Fire Left & 1 \\ Brake Fire Right & 2 \\ Come Close & 3 \\ Cut Engine Left & 4 \\ Cut Engine Right & 5 \\ Down & 6 \\ Engine Start Left & 7 \\ Engine Start Right & 8 \\ Follow & 9 \\ Left & 10 \\ Move Away & 11 \\ Negative & 12 \\ Release Brake & 13 \\ Right & 14 \\ Slow Down & 15 \\ Stop & 16 \\ Straight & 17 \\ Take Photo & 18 \\ Up & 19 \\ Null Class & 20 \\ 
\bottomrule \end{tabularx} 
    \vspace{-4mm}
\end{table}

\subsection{Data Collection}
The dataset was collected from eleven participants, comprising seven males and four females. They received instructions and demonstrations of the defined gestures, with natural variations permitted. Each participant completed five sessions. In each session, they performed 80 gesture instances, consisting of four randomized repetitions of each gesture. The recorded sensor channels from each device are summarized in \cref{tab:glove_channels} and \cref{tab:watch_channels}. 
To synchronize modalities, five claps at the beginning and end of each session were used as reference points between the video and sensor streams. 
A temporal scaling factor—defined as the ratio between the video sequence length and the sensor sequence length—was applied to compress or stretch sensor data to match video duration.
All gesture segments were labeled using LabelStudio \cite{LabelStudio}.

Several preprocessing steps were applied to the raw sensor data before training the gesture recognition model. Sensor channels from the accelerometer, gyroscope, quaternion orientations, and capacitive sensors were used for model training. IMU data were obtained from the Apple Watch to ensure alignment with the quaternion orientations. Capacitive sensors from the gloves were included as additional inputs, while the glove IMU data served only as a complement to the Apple Watch IMU. Standardization was applied to each sensor device’s raw data to mitigate inter-device variability and measurement noise. The sensor data were then segmented using a sliding-window approach, where each window corresponded to a fixed-length segment of sensor readings. A window size of 3.0 seconds was used to capture sufficient temporal context for gesture recognition, with a step size of 1.0 seconds to maintain continuity between windows. Each window was labeled by a threshold-voting mechanism, which determined its label based on the proportion of overlap between the window and the fully annotated gesture sequence. A threshold of 0.75 was employed, and windows that did not meet the threshold for any class were labeled as \texttt{Null}. Windows labeled as \texttt{Null} were excluded, ensuring that the model was trained exclusively on valid gestures. All windows containing claps for synchronization were also removed before training. 

\subsection{Training Details}

Training was conducted using both leave-one-session-out(LOSO) and leave-one-participant-out(LOPO) strategies across five sessions and six participants. The dataset splits were generated from the remaining six participants after excluding all sessions identified as corrupted and unsuitable for training. Models were trained using cross-entropy loss for multi-class gesture classification. Training was performed with the Adam optimizer \cite{kingma2014adam}, using a learning rate of $1\times10^{-4}$ and beta values of $(0.9, 0.999)$. Models were trained for a maximum of 150 epochs with a batch size of 32 on a single NVIDIA GeForce RTX 4090 24GB GPU. Early stopping was applied based on the F1-score of the validation set. Model performance on gesture recognition was evaluated using precision, recall, and macro F1-score. The proposed multimodal dataset and associated code for preprocessing and model implementation will be released publicly as supplementary materials upon acceptance.

\begin{table}[t] 
\caption{Glove Sensor Channels} \label{tab:glove_channels}
\centering 
\begin{tabular}{l|l}
\toprule
\textbf{Sensor Type} & \textbf{Feature Names} \\
\midrule 
Capacitive Sensors (CAPA) & CAPA\_0, CAPA\_1, CAPA\_2, CAPA\_3\\
Accelerometer (ACC) & ACC\_X, ACC\_Y, ACC\_Z \\ 
Gyroscope (GYRO) & GYRO\_X, GYRO\_Y, GYRO\_Z \\ 
\bottomrule 
\end{tabular} 

\end{table}

\begin{table}[t]
\caption{Apple Watch Sensor Channels} \label{tab:watch_channels}
\centering
\resizebox{\linewidth}{!}{
\begin{tabular}{l|l}
\toprule
\textbf{Sensor Type} & \textbf{Feature Names} \\
\midrule 
Accelerometer (ACC) & ACC\_X, ACC\_Y, ACC\_Z \\
Gyroscope (GYRO) & GYRO\_X, GYRO\_ GYRO\_Z \\
Quaternion Orientation (QUAT) & QUAT\_W, QUAT\_X, QUAT\_Y, QUAT\_Z \\
\bottomrule
\end{tabular}
}
\vspace{-3mm}
\end{table}

\section{Results}

\subsection{Quantitative Evaluation with Comparisons}

\begin{table}[!t]
\centering
\caption{Model Performance Comparison}
\label{tab:performance}
\begin{tabularx}{\columnwidth}{X c c c}
\toprule
\textbf{Model} & \textbf{F1 Score (\%)} & \textbf{Precision (\%)} & \textbf{Recall (\%)} \\
\midrule
\multicolumn{4}{l}{\textit{(LOSO Split)}} \\
\addlinespace[0.5em]
LLR Fusion \newline (sensor) & \textbf{95.40 $\pm$ 5.91} & \textbf{95.99 $\pm$ 4.84} & \textbf{95.52 $\pm$ 5.85} \\
\addlinespace[0.3em]
Self-Attention \newline Fusion (sensor) & 94.42 $\pm$ 5.20 & 95.05 $\pm$ 4.44 & 94.45 $\pm$ 5.09 \\
\addlinespace[0.5em]
PoseConv3D (video) & 94.70 $\pm$ 2.33 & 95.03 $\pm$ 1.93 & 94.73 $\pm$ 2.35 \\
\addlinespace[0.5em]
\midrule
\addlinespace[0.5em]
\multicolumn{4}{l}{\textit{(LOPO Split)}} \\
\addlinespace[0.5em]
LLR Fusion \newline  (sensor) & \textbf{93.59 $\pm$ 5.20} & 94.45 $\pm$ 4.70 & 93.59 $\pm$ 5.31 \\
\addlinespace[0.5em]
Self-Attention \newline Fusion (sensor) & 92.97 $\pm$ 6.11 & 94.40 $\pm$ 4.73 & 93.10 $\pm$ 5.98 \\
\addlinespace[0.5em]
PoseConv3D (video) & 93.39 $\pm$ 5.44 & \textbf{95.09 $\pm$ 3.40} & \textbf{93.75 $\pm$ 4.91} \\
\bottomrule
\end{tabularx}
\end{table}

\begin{figure}[!t]
    \centering
    \includegraphics[width=\columnwidth]{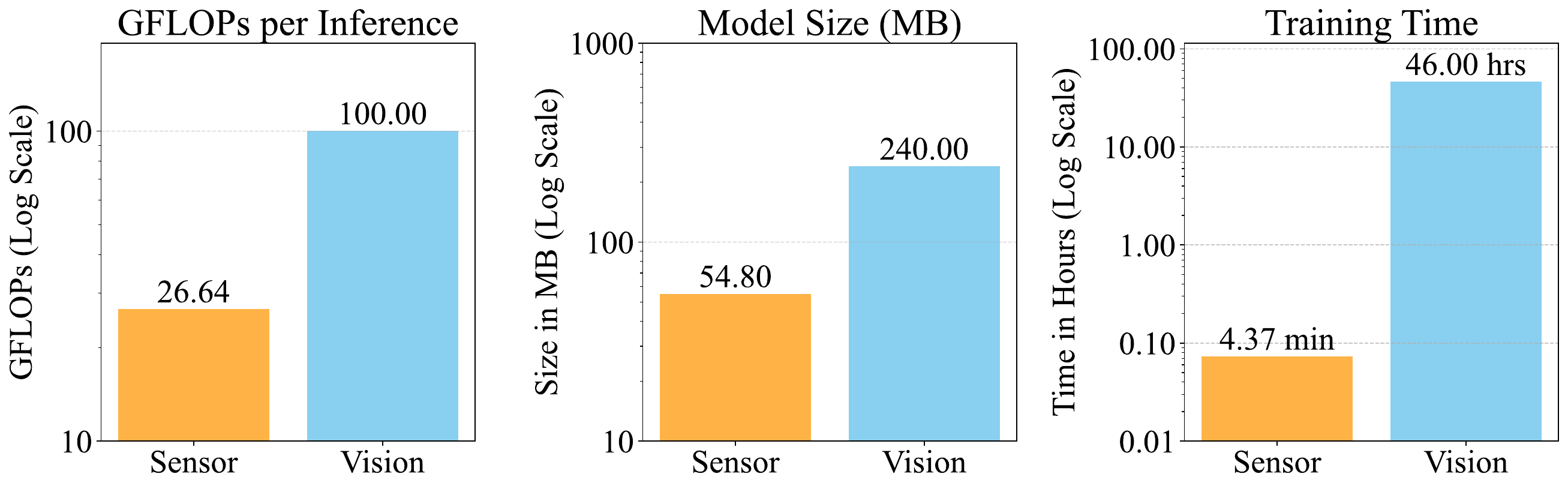}
    \caption
    {Computational resource comparison between sensor-based (orange) and 
         vision-based (blue) models. From left to right, the plots compare 
         GFLOPs per inference, model size, and training time. 
         Training time is reported as the average across all dataset split variations.}
         \vspace{-5mm}
    \label{fig:computation}
\end{figure}

We evaluate two fusion methods in our sensor-based model and a single vision-based approach (PoseConv3D) using F1-score, precision, and recall, with results shown in \cref{tab:performance}. Under the LOSO split, the sensor-based LLR fusion model achieved the highest scores across all metrics. Under the LOPO split, it outperformed PoseConv3D in F1-score, while showing comparable precision and recall with only minor differences. Overall, metrics in the LOPO split were lower than in the LOSO split, suggesting that LOPO is a more challenging task. In addition, the sensor-based model was computationally more efficient, requiring fewer GFLOPs per inference, a smaller model size, and shorter training time, as shown in \cref{fig:computation}. For resource usage comparison, the LLR fusion was used to represent the sensor-based approach, while PoseConv3D represented the vision-based approach.

\subsection{Qualitative Evaluation for Interpretability}

\begin{figure}[!t] 
    \centering 
    \begin{subfigure}[b]{0.48\columnwidth} 
        \centering
        \includegraphics[width=\linewidth,height=3.5cm]{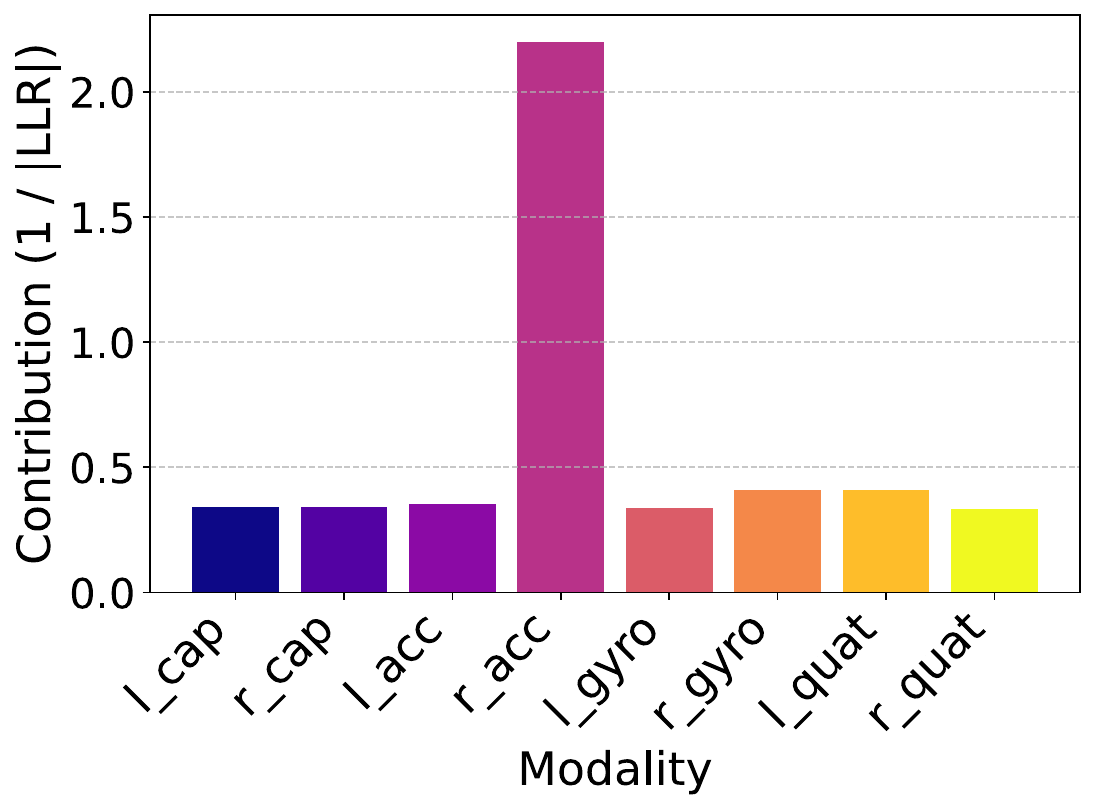} 
    \end{subfigure}
    \hfill
    \begin{subfigure}[b]{0.48\columnwidth} 
        \centering
        \includegraphics[width=\linewidth,height=3.5cm]{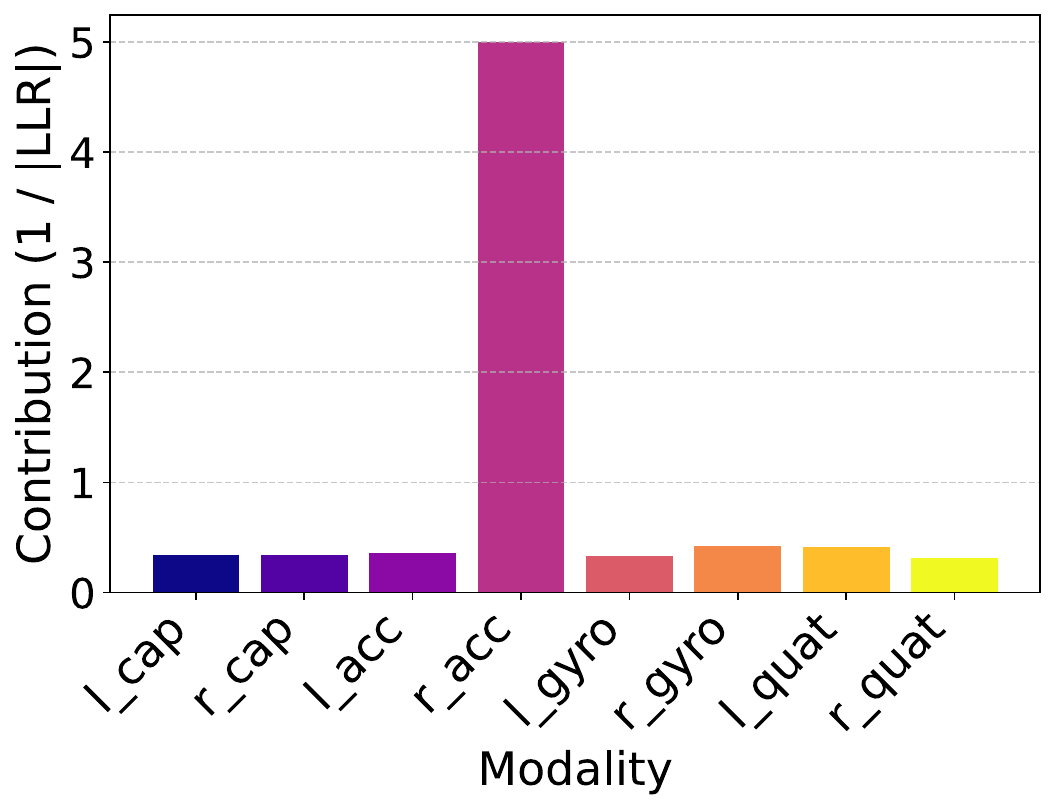}
    \end{subfigure}
    \caption{Randomly sampled LLR contribution per modality, when correctly predicted $\texttt{Move Away}$ gesture}
    \label{fig:moveaway_llr}
    \vspace{-3mm}
\end{figure}

\begin{figure}[!t] 
    \centering 
    \begin{subfigure}[b]{0.48\columnwidth} 
        \centering
        \includegraphics[width=\linewidth,height=3.5cm]{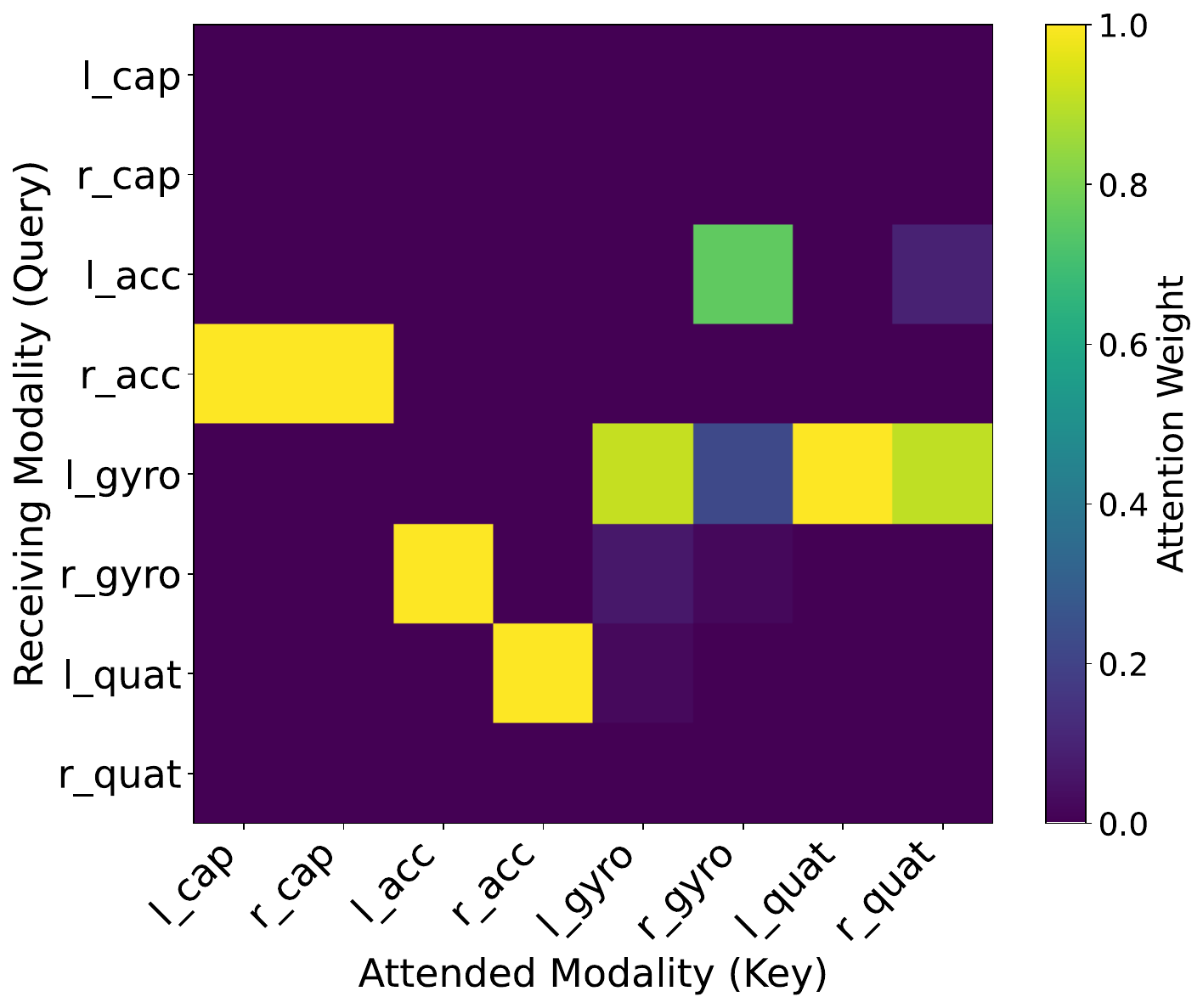} 
    \end{subfigure}
    \hfill
    \begin{subfigure}[b]{0.48\columnwidth} 
        \centering
        \includegraphics[width=\linewidth,height=3.5cm]{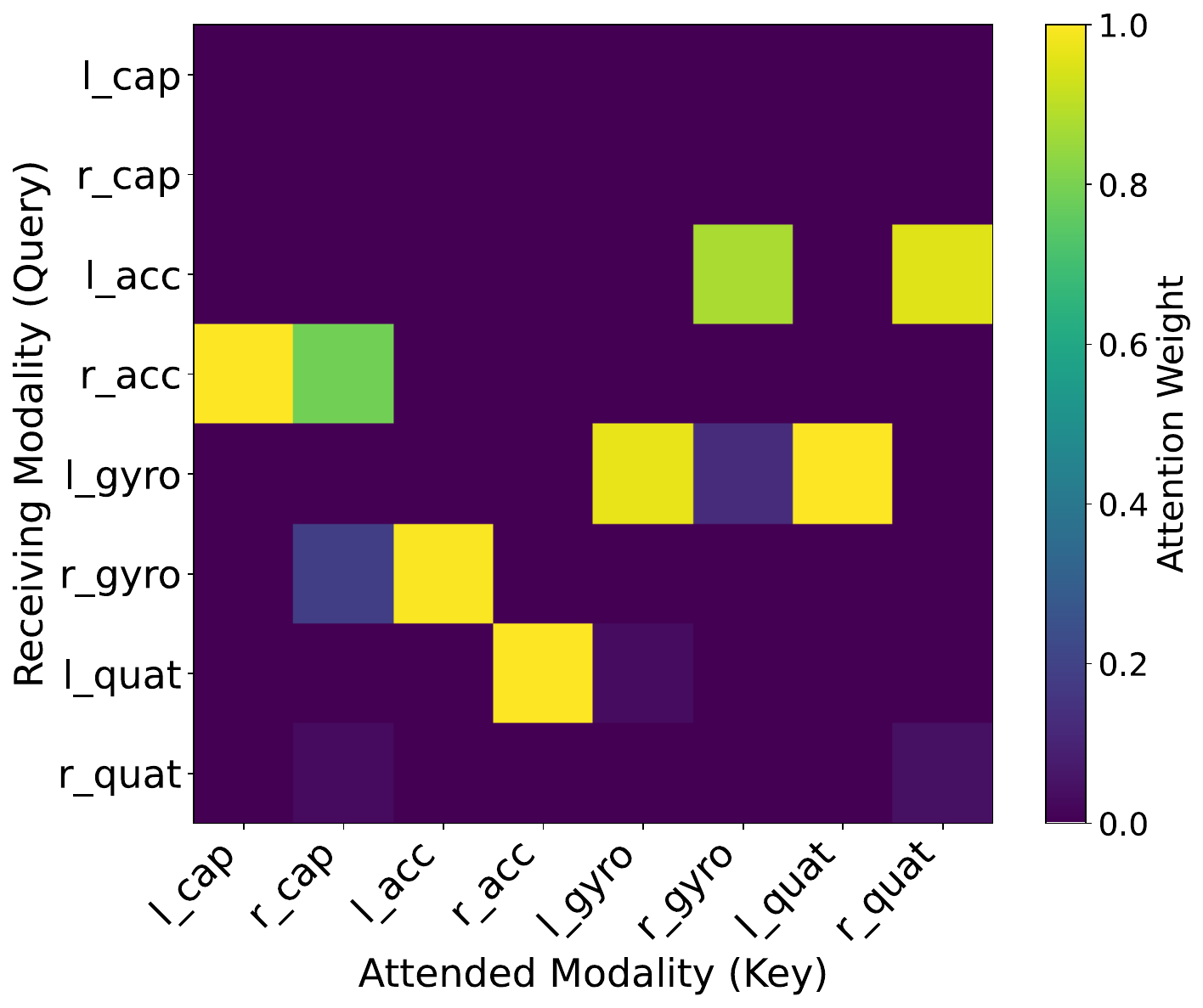}
    \end{subfigure}
    \caption{Randomly sampled attention weights for a case when correctly predicted $\texttt{Move Away}$ gesture}
    \vspace{-5mm}
    \label{fig:moveaway_attn_w}
\end{figure}

Model interpretability was assessed using models trained on the LOPO split. From each fusion method, LLR values per modality and attention weights were randomly sampled during inference. \cref{fig:moveaway_llr} and \cref{fig:moveaway_attn_w} illustrate these values for correct predictions of the class \texttt{Move Away}. LLR values are transformed to the inverse of their magnitude so that contribution magnitudes are represented as positive rather than negative values. As shown in \cref{fig:gesture}, this gesture involves waving the right hand horizontally with the palm facing forward. The predominantly linear motion of this gesture led to the hypothesis that accelerometer features would be especially relevant for correct classification. \cref{fig:moveaway_llr} shows that the right-hand accelerometer feature representation exhibited the highest LLR magnitude, indicating its strong contribution to correctly classifying the gesture as \texttt{Move Away}. All other modalities showed visibly lower magnitudes than that of the right-hand accelerometer. The self-attention fusion mechanism provides a more granular insight that complements the LLR findings. As shown in \cref{fig:moveaway_attn_w}, the attention patterns also highlighted the right-hand accelerometer as a key component. The highest attention weights were associated with the right-hand accelerometer, whose representation attended most strongly to the capacitive sensor features, while also being the main target of attention from the left-quaternion orientation feature. While these patterns suggest that the model is learning relevant inter-feature relationships, we treat them as diagnostic, as high attention does not guarantee a causal role in the final prediction \cite{serrano2019attention}.

\subsection{Ablation Study}

\begin{table}[!t]
\centering
\caption{Performance Comparison in Modality Combinations in Terms of F1 Score (LOSO vs. LOPO)}
\label{tab:ablation_combined}
\renewcommand{\arraystretch}{1.2}
\begin{tabularx}{\columnwidth}{X c c}
\toprule
\multirow{ 2}{*}{\textbf{Sensor Modalities}} & \multicolumn{2}{c}{\textbf{F1 Score (\%)}} \\
& \textbf{LOSO} & \textbf{LOPO} \\
\midrule
\multicolumn{3}{l}{\textit{(All 4 sensors)}} \\
\addlinespace[0.5em]
CAP + ACC + GYRO + QUAT & 95.40 $\pm$ 5.91 & 93.59 $\pm$ 5.20 \\
\addlinespace[0.5em]
\midrule
\multicolumn{3}{l}{\textit{(3 sensors)}} \\
\addlinespace[0.5em]
CAP + ACC + GYRO & 95.80 $\pm$ 5.61 & 91.78 $\pm$ 4.11 \\
CAP + ACC + QUAT & 95.24 $\pm$ 5.40 & 92.11 $\pm$ 4.52 \\
ACC + GYRO + QUAT  & 95.75 $\pm$ 5.39 & \textbf{93.75} $\pm$ \textbf{4.38} \\
\addlinespace[0.5em]
\midrule
\multicolumn{3}{l}{\textit{(2 sensors)}} \\
\addlinespace[0.5em]
ACC + QUAT  & 95.30 $\pm$ 5.54 & 92.36 $\pm$ 4.92 \\
ACC + GYRO  & \textbf{95.81} $\pm$ \textbf{5.05} & 92.00 $\pm$ 4.25 \\
CAP + QUAT  & 93.88 $\pm$ 5.51 & 87.96 $\pm$ 7.23 \\
CAP + ACC   & 94.66 $\pm$ 5.91 & 89.82 $\pm$ 5.01 \\
\addlinespace[0.5em]
\midrule
\multicolumn{3}{l}{\textit{(1 sensor)}} \\
\addlinespace[0.5em]
CAP & 49.22 $\pm$ 8.99 & 7.98 $\pm$ 4.27 \\
ACC & 94.51 $\pm$ 5.89 & 90.05 $\pm$ 5.96 \\
GYRO & 95.21 $\pm$ 5.42 & 90.05 $\pm$ 6.70 \\
\addlinespace[0.5em]
\bottomrule
\end{tabularx}
    \vspace{-5mm}
\end{table}

We conducted an ablation study to evaluate the impact of individual sensor modalities on model performance, quantified by the F1 score. This analysis employed our LLR fusion-based sensor approach and was evaluated under both LOSO and LOPO cross-validation splits.

In the LOSO splits, we observed minimal F1 score degradation across many sensor combinations, even when reducing the sensor count from four to two. This suggests that the gesture recognition task within the LOSO setting may be sufficiently simple, allowing models with fewer sensors to achieve performance comparable to those using the full sensor suite.

Conversely, the performance disparities between different sensor combinations were significantly more pronounced in the LOPO splits. This indicates that the LOPO evaluation presents a more challenging generalization task, where a reduced sensor count is often insufficient to match the performance of a richer sensor set. Notably, in the LOPO split, the top-performing combination resulted from excluding the capacitive sensor. In contrast, the same combination of capacitive sensors, accelerometer, and gyroscope ranked third in the LOSO split, but its F1 score was only marginally different from the other top-five combinations. These results suggest the LOSO split lacks the sensitivity to robustly differentiate the efficacy of various sensor combinations.

Further analysis of the capacitive sensor modality reinforces this finding. The model relying solely on capacitive sensors yielded significantly lower F1 scores than any other single-sensor or multi-sensor combination in both validation splits, with a particularly substantial performance degradation in the LOPO split. We hypothesize that this is due to our defined gesture set, which predominantly involves large-scale arm and hand motions rather than the fine-grained, individuated finger movements that capacitive sensors are designed to detect.

This hypothesis is substantiated by the observation that the capacitive-only model collapsed its predictions to the \texttt{Take Photo} gesture. This specific gesture requires participants to "raise both hands in front of the face and form a rectangular frame by extending the index fingers vertically and the thumbs horizontally," a motion that uniquely engages the fingers in a static, articulated pose. The model's comprehensive failure to recognize other classes, which lack this intensive finger engagement, underscores the capacitive sensor's limited utility for our target gesture vocabulary.

    
    
  

\subsection{Discussion}
Our LLR-based multimodal framework addresses key limitations of gesture-driven teleoperation in hazardous, real-world deployments. By integrating heterogeneous wearable modalities into an interpretable and efficient model, the system enables reliable control in safety-critical scenarios such as disaster response and industrial maintenance, where understanding sensor-level contributions is essential for diagnosing failures and building operator trust. Unlike vision-based systems that degrade under smoke, low light, or occlusion, inertial and capacitive sensing remain robust in such environments. The gesture vocabulary, inspired by aircraft marshalling \cite{choi2008visual}, further supports intuitive adoption in emergency operations without extensive retraining.
Compared with vision-based baselines, the sensor-based framework reduces computational cost and model size, enabling real-time deployment on resource-constrained edge devices without reliance on cloud infrastructure, and extending battery life by lowering computational load and minimizing operational interruptions \cite{romero2024cellular}.

Despite strong performance, several limitations should be noted. Attention weights were used for model explainability, but they can be misleading \cite{serrano2019attention}. To mitigate this, LLR values were incorporated to capture direct modality-level contributions, preserving a clear relationship between features and final predictions \cite{ribeiro2016should}. Additionally, several recording sessions were lost due to hardware issues, and all data were collected indoors, leaving robustness in uncontrolled environments uncertain. Future work should expand data collection to more diverse operational conditions, including outdoor, dynamic, and cluttered environments.
The current implementation processes modality features sequentially, introducing avoidable latency during training and inference. Future work will adopt a parallel architecture in which unimodal encoders operate simultaneously, reducing inference delay and improving scalability for real-time robotic deployment.

Another limitation concerns the scope of the gesture set. While its foundation in aircraft marshalling protocols \cite{choi2008visual} is a key strength for standardization, this choice also means it requires further operation-specific adjustment to achieve universal applicability. This protocol is highly optimized for vehicle navigation (e.g., \texttt{Stop}, \texttt{Left}, \texttt{Right}), but it lacks the necessary vocabulary for the complex manipulation or instrumentation commands required in other domains. For broader adoption, the mapping between gestures and control commands must be adapted or extended. Future work should incorporate new, domain-specific gestures for tasks such as industrial inspection or emergency rescue.

Relatedly, the physical properties of our current gesture set introduced a sensor-specific limitation. As shown in our ablation study, the capacitive sensor's contribution was minimal, which we attribute to the vocabulary’s emphasis on large-scale arm motions rather than fine finger articulation. Although they did not yield significant accuracy gains overall, the glove sensors provided complementary information in cases where wrist-based inertial signals were ambiguous. This limited improvement reflects the aircraft marshalling–inspired design, which prioritizes gross limb movements. Nonetheless, their inclusion highlights the framework’s potential for future gesture dictionaries that involve finer finger-based interactions, where capacitive sensing is expected to play a more significant role. Expanding the gesture vocabulary to include actions requiring precise finger control—such as pinches, taps, or other subtle manipulations—could therefore unlock a substantial performance boost, with capacitive sensing providing unique and critical signals that enhance the model’s discriminative power. Releasing this multimodal dataset further enables future studies to explore such extensions and develop sensor fusion strategies that more effectively combine wrist and glove modalities.

\section{Conclusions}

In this work, we presented a multimodal gesture recognition framework for mobile robots and drones that integrates inertial signals from Apple Watches with capacitive sensing from textile gloves. Using a log-likelihood ratio (LLR)–based late fusion strategy, our method enhances recognition performance while providing interpretability through modality-specific contributions. We also introduced a dataset of 20 aircraft marshalling–inspired gestures with synchronized RGB, IMU, and capacitive data. Experiments showed that our approach outperforms vision-based baselines in F1 score while requiring lower computational cost and a smaller model size.

Future work will expand data collection to larger and more diverse settings, including continuous gesture streams, multi-robot coordination, and operational conditions beyond controlled indoor settings to further assess robustness. Beyond drone teleoperation, the proposed approach may extend to broader human–robot interaction scenarios where robustness, efficiency, and interpretability are critical for safe operation. This work advances practical, interpretable, and sensor-based gesture recognition for intuitive and reliable robot control in real-world environments.


\section*{ACKNOWLEDGMENT}

This work was supported by the IITP grant funded by the MSIT (RS-2019-II190079, AI Graduate School Program (Korea University) and IITP-2026-RS-2024-00436857, IITP-ITRC), and by the Technology Innovation Program (RS-2025-25453819) funded by the MOTIR. This work was also supported by the BMBF in the project CrossAct (01IW25001). 



\bibliographystyle{IEEEtran}
\bibliography{refs}

\end{document}